%% file: V2.tex
\documentclass[conference]{IEEEtran}
\IEEEoverridecommandlockouts
\usepackage{cite}
\usepackage{amsmath,amssymb,amsfonts}
\usepackage{algorithmic}
\usepackage{graphicx}
\usepackage{textcomp}
\usepackage{xcolor}
\usepackage{float}
\usepackage{url}

\usepackage{mathptmx}
\usepackage{cite,amsmath,amssymb,graphicx,booktabs,url}
\usepackage{tikz,pgfplots,balance}
\usetikzlibrary{arrows.meta,positioning,fit,backgrounds,calc}
\usepgfplotslibrary{fillbetween,groupplots}
\pgfplotsset{compat=1.18}
\newcommand{\method}{RSSM--RLPD}
\newcommand{\grumethod}{GRU--RLPD}
\newcommand{\vect}[1]{\boldsymbol{#1}}

\newcommand{\HWGoal}{100}
\newcommand{\HWBaseline}{0}

\definecolor{orlblue}{HTML}{27628D}
\definecolor{orlpurple}{HTML}{6B5CA5}
\definecolor{orlamber}{HTML}{C47A24}
\definecolor{orlgreen}{HTML}{35866B}
\definecolor{orlgray}{HTML}{52606D}

\def\BibTeX{{\rm B\kern-.05em{\sc i\kern-.025em b}\kern-.08em
    T\kern-.1667em\lower.7ex\hbox{E}\kern-.125emX}}
\begin{document}

\title{Reinforcement Learning from Real-Robot Interaction without High-Quality State Estimates}

\author{
Ryan Gao$^{*}$,
Ashvin Anilkumar$^{*}$,
Nicholas Corrado,
William Cong,
Josiah Hanna
\\[0.2em]
\small University of Wisconsin--Madison
\\
\small
\texttt{\{rgao97, anilkumar4, ncorrado, wycong, jphanna\}@wisc.edu}
\\[0.2em]
\small $^{*}$Equal contribution.
}
\maketitle

\begin{abstract}

A longstanding goal for robot reinforcement learning (RL) is that robots learn from their own physical experience, as humans and animals do. A key obstacle to learning directly from physical experience, especially outside instrumented lab environments, is that the state of the world is usually unobservable to the learner and must be estimated from on-board sensing. Errors in that estimate corrupt both the policy input and potentially the reward signal. We study this challenge using the contact-rich task of a bipedal robot learning to score soccer goals. We estimate all key state variables on-board with an unscented Kalman filter (UKF) and no external instrumentation (e.g., motion capture or overhead cameras). Because the UKF may incorrectly estimate state when the ball is unobserved, we learn a new state representation with a recurrent state-space model (RSSM) that, critically, only corrects the RSSM's latent state when the ball is observed. Building on RLPD \cite{ball2023rlpd}, we introduce RSSM-RLPD and apply it to the task of a NAO robot learning to dribble a soccer ball into a goal.
We show that the robot learns to score from multiple starting positions after 30 minutes of real-world interaction, reaching \HWGoal\% success against \HWBaseline\% for a feed-forward policy under the same training budget. We complement our physical robot experiments with a simulation study and show that RSSM-RLPD learns policies that achieve a higher success rate than behavior cloning, a feed-forward, and GRU baseline. Taken together, our results show that the addition of a learned state representation to existing on-robot learning methods enables RL from real-robot data even with a high degree of partial observability.
%We conclude that estimation error, not data volume, is the binding constraint on learning from physical experience in this setting.
\end{abstract}

\section{Introduction}
\label{sec:introduction}

Robots that learn from their own physical experience, as humans and animals do, have been a longstanding goal of robot learning. Recent systems have made real-world reinforcement learning (RL) practical enough to acquire useful skills in hours (or even minutes) rather than weeks, but they have largely been applied to tasks with high state observability, e.g., quadruped locomotion \cite{smith2023walkinthepark} or manipulation tasks with a camera mounted over the workspace \cite{luo2024serl,ball2023rlpd,wagenmaker2025steering}, or external instrumentation such as overhead cameras \cite{wu2023daydreamer}. Outside of such settings, applying RL requires a robot to estimate the state of the world from its own sensors, and such estimates are often noisy and inaccurate. Objects leave the field of view, detections drop out, and the robot's own motion perturbs the estimate. The learner therefore has only an uncertain and partial view of its true state. 
% Our robot processes onboard camera images into ball and landmark detections. Its estimation stack uses odometry, an unscented Kalman filter (UKF) for robot localization \cite{laue2009efficient}, and a Kalman-filter ball tracker. The learner receives features derived from these estimates.

We study the challenge of learning with uncertain state estimates on a walking robot learning to dribble a ball into a goal. 
The ball can leave the robot's field of view during approach and contact. Each attempt also costs real time and often requires ball replacement or operator intervention, so a practical method must be data efficient as well as tolerant of an uncertain state estimate. 
Our robot processes onboard camera images into ball and landmark detections. Its estimation stack uses odometry, an unscented Kalman filter (UKF) for robot localization \cite{laue2009efficient}, and a Kalman-filter ball tracker. The learner receives features derived from these estimates.
To enable data-efficient RL, we build on SERL \cite{luo2024serl} and RLPD \cite{ball2023rlpd}, which mix scarce online experience with demonstration data.
% We formulate the task as a partially observable Markov decision process (POMDP): past ball motion and contact can inform actions when current observations are incomplete \cite{kaelbling1998pomdp}.

\begin{figure}[t]
    \centering
    \includegraphics[width=0.7\linewidth]{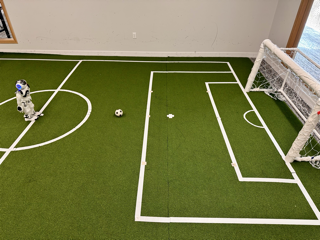}
    \caption{Physical setup: a NAOv6 humanoid robot on the $7\times4.8$~m half-field, learning to dribble the ball into the goal from real experience and on-board sensing. RL is challenged by on-board sensing, which often yields inaccurate estimates of key state information, such as the ball’s location when it is outside the field of view.}
    \label{fig:goal_setup}
\end{figure}

Even with data-efficient RL algorithms and demonstrations, state estimate quality is too poor to simply apply these algorithms on top of the state representation output by the UKF.
We address this challenge by introducing \method{}, which learns a new state representation using a recurrent state-space model (RSSM). \method{} maintains a latent belief over the task state with the RSSM, running on top of the robot's existing state estimator. When a fresh ball observation arrives, \method{} corrects the belief with a posterior update; when one does not, \method{} advances the belief with its own predictive prior. The actor and critics act on the resulting belief statistics together with an observation-gated innovation feature. We train the representation with two signals: a predictive loss that encourages it to retain information about future observations and interaction outcomes, and critic gradients to make it useful for the task.
% \method{} adds memory to the learning loop rather than replacing any part of it, leaving the robot's state estimator and the recorded training rewards untouched, so that past evidence remains available for control and learning when current measurements are insufficient.

% We evaluate our system on the task of a robot learning to score from different soccer ball starting positions.
We evaluate \method{} on hardware and in simulation. On hardware, the robot learns to score the first goal in 10 minutes and to score reliably from multiple positions within 30 minutes of on-board experience with no external instrumentation. A simulation study shows that the learned belief, rather than the added capacity or the demonstration data, accounts for the improvement. Together, these results indicate that maintaining a predictive belief over an imperfect on-board estimate can make learning from physical experience tractable in this setting.

\section{Related Work}
\label{sec:related-work}

\subsection{Sample-Efficient On-Robot Learning}
Within the current era of deep RL, there have been many notable success stories of applying RL on physical hardware.
Smith et al.\ demonstrate model-free RL for quadruped locomotion, learning forward walking in 20 minutes on physical hardware
\cite{smith2023walkinthepark}.
RLPD combines demonstration data and online replay \cite{ball2023rlpd};
SERL and HIL-SERL integrate efficient learning with robot control and
human interventions \cite{luo2024serl,luo2025hilserl}.
Wagenmaker et al.\ improve manipulation policies through real-world
interaction by applying RL in the latent-noise space of a pretrained
diffusion policy \cite{wagenmaker2025steering}.
SAC and REDQ underpin the actor--critic and replay-update design
\cite{haarnoja2018sac,chen2021redq}. On-robot POMDP learning includes
Bayesian planning with model priors \cite{nguyen2023onrobot} and
recurrent, demonstration-assisted assembly \cite{nguyen2024softwrist}.
In contrast to these works, our work studies ball contact under intermittent visual observations.

World models have emerged as a powerful framework for improving learning and decision-making in robotic systems. PlaNet plans with latent dynamics \cite{hafner2019planet}; Dreamer
learns from latent imagination \cite{hafner2020dreamer}.
DayDreamer learns world models from physical interaction and trains
behavior through imagined latent trajectories for locomotion,
manipulation, and navigation \cite{wu2023daydreamer}.
In contrast, our approach uses the world model to provide predictive features that support actor--critic learning from recorded interactions.
%Our work extends this sample-efficient learning approach to sustained ball pushing, using predictive memory to support control and learning from intermittent, estimator-derived observations.

\subsection{Memory and Predictive Representations}
Recurrent model-free algorithms are strong baselines for POMDPs
\cite{ni2022recurrent}. Predictive supervision helps train such memory:
DVRL jointly learns a generative model, inference, and policy
\cite{igl2018dvrl}; SLAC and VRM combine latent models with off-policy
actor--critic learning \cite{lee2020slac,han2020vrm}. Belief-Grounded
Networks use beliefs computed in simulation to supervise history
features \cite{nguyen2021bgn}.
Recurrent Kalman Networks structure inference around prediction and
available measurements \cite{becker2019rkn}. We use the different
availability of ball and robot information: current robot-to-goal
context supports prediction, while fresh ball evidence determines
correction. Predictive and critic objectives train this representation
from demonstration and online replay. 

\subsection{Robot Soccer}

Simulation-trained policies achieve agile bipedal soccer
\cite{haarnoja2024soccer} and active perception through egocentric vision
and recurrence \cite{tirumala2025soccer}. Wang et al. combine structured
ball and goal observations with history encoding and auxiliary state
reconstruction \cite{wang2026vision}.
Labiosa et al. integrate learned behaviors with classical perception,
localization, and walking modules \cite{labiosa2025soccer}. In our study, we build
on this modular setting to learn goal-directed ball dribbing from
limited physical interaction using predictive memory over estimator
outputs.

\section{Task and System}
\label{sec:task_system}

\subsection{Task Formulation}

We formulate the task of dribbling a soccer ball into a goal as a \textit{partially observable Markov decision process} (POMDP) \cite{kaelbling1998pomdp} defined by the tuple $(\mathcal S,\mathcal A,\mathcal O,T,\Omega,R,d_0,\gamma)$ where $\mathcal S$ is the state space, $\mathcal A$ is the action space, $\mathcal O$ is the observation space, $T$ is the transition function, $\Omega$ is the observation function, $R$ is the reward function, $d_0$ is the initial state distribution, and $\gamma\in[0,1)$ is the discount factor.

At each step $t$, the agent receives a partial observation $o_t$ of the underlying state $s_t$ according to the observation function $\Omega$ and then selects an action $a_t$ from a policy $\pi$ conditioned on a function of the observation-action history $h_t=f(o_{0:t},a_{0:t-1})$. The environment then transitions to the next state according to the transition function $T$, and the agent receives a scalar reward $r_t$. The goal is to learn a policy $\pi^\ast$ that maximizes the discounted return
\begin{equation}
 J(\pi)=\mathbb E_\pi\!\left[\sum_{t=0}^{\infty}\gamma^t r_t\right],
 \label{eq:pomdp_objective}
\end{equation}
where $s_0\sim d_0$ and the expectation is taken with respect to trajectories $s_0,a_0,s_1,a_1,\dots$ induced by the policy $\pi$. 

The following subsections discuss how observations, actions, rewards, and episode terminations are realized on the physical robot. The simulated environment in Section~\ref{sec:experiments} implements a similar interface with differences noted in Section~\ref{sec:sim_setup}.

\subsection{Robot Platform}

We use a NAOv6 humanoid robot equipped with on-board perception, state estimation, and locomotion modules. The actor and learner run on an external workstation: the robot wirelessly streams observations to the workstation, which returns actions to the robot while asynchronously updating the policy. Figure~\ref{fig:on_robot_loop} illustrates this interaction loop. No additional instrumentation is used; all observations available to the learner are derived exclusively from the robot's on-board state estimator.

\input{figures/on_robot_loop_inline.tex}

\subsection{Observations}
\label{sec:task_observation}

The on-board state estimator provides an estimate of the robot pose $(\hat x_t^r,\hat y_t^r,\hat\theta_t^r)$ and ball position $(\hat x_t^b,\hat y_t^b)$ in field coordinates, along with a ball visibility flag $v_t$ and the age of the most recent ball observation $\tau_t$.
\begin{equation}
 \vect{m}_t=[\hat x_t^r,\hat y_t^r,\hat\theta_t^r,\hat x_t^b,\hat y_t^b,v_t,\tau_t]
 \label{eq:estimator_interface}
\end{equation}
From $\vect{m}_t$, we construct a 12-dimensional policy observation
\begin{equation}
\begin{aligned}
\vect{o}_t
&= [\vect{o}^{\text{ball}}_t,
\vect{o}^{\text{robot}}_t],\\
\vect{o}^{\text{ball}}_t
&= [\tilde x_t^{b,r},\tilde y_t^{b,r},v_t,\tau_t,
\rho_t,u_t,\tilde{\dot x}_t^{b,r},\tilde{\dot y}_t^{b,r}],\\
\vect{o}^{\text{robot}}_t
&= [d_{\parallel,t}^{r,g},d_{\perp,t}^{r,g},
\sin\theta_t^{r,g},\cos\theta_t^{r,g}].
\end{aligned}
\label{eq:policy_observation}
\end{equation}
which we decompose into ball-related quantities $\vect{o}^{\text{ball}}_t$ and robot-related quantities $\vect{o}^{\text{robot}}_t$. 
Here, $(\tilde x_t^{b,r},\tilde y_t^{b,r})$ denotes the estimated ball position in the robot frame, while $(\tilde{\dot x}_t^{b,r},\tilde{\dot y}_t^{b,r})$ denotes the estimated ball velocity, computed from consecutive ball position estimates. When the ball is not visible, its position estimate is replaced by a sentinel value and its velocity is set to zero.
The quantities $\rho_t$ and $u_t$ measure the reliability of the ball-state estimates, with $\rho_t=1$ if $v_t=1$ and $\rho_t=\exp[-\tau_t/(0.5,\mathrm{s})]$ otherwise; $u_t$ is a log-scaled uncertainty proxy.

Finally, $d_{\parallel,t}^{r,g}$ and $d_{\perp,t}^{r,g}$ denote the robot's signed distances from the goal along and perpendicular to the goal's longitudinal axis, respectively, while $(\sin\theta_t^{r,g}, \cos\theta_t^{r,g})$ encodes the robot's corresponding bearing relative to the goal. All methods discussed in Section~\ref{sec:experiments} receive this same policy observation.

\subsection{Actions}
\label{sec:task_action}

The policy outputs a three-dimensional action containing the desired forward, lateral, and rotational body velocities at a nominal control frequency of 30~Hz.
\begin{equation}
 \vect{a}_t=[a_t^{\mathrm{fwd}},a_t^{\mathrm{lat}},a_t^{\mathrm{rot}}]^\top
 \in[-1,1]^3
 \label{eq:continuous_action}
\end{equation}
The resulting action is then post-processed by clipping $a_t^{\mathrm{rot}}$ to a maximum magnitude of $0.6$, smoothing against the previous executed action ($\vect{a}_t=0.35\,\vect{a}_{t-1}^{\mathrm{exec}}+0.65\,\vect{a}_t$), and then applying a safety filter that suppresses movement toward the field boundaries and goal keep-out boundaries. The resulting action $\vect{a}_t^{\mathrm{exec}}$ is sent to the locomotion controller and recorded as the executed action.

\subsection{Reward}
\label{sec:reward}

We compute the task reward from the on-board state estimates $\vect{m}_t$. To prevent unreliable ball estimates from producing spurious shaping rewards, we apply the progress terms only when the ball is freshly tracked. Specifically, let $\chi_t=1$ when the ball is continuously observed throughout the transition without loss or reacquisition, and $\chi_t=0$ otherwise. The online reward is
\begin{equation}
\begin{aligned}
r_t=&\chi_t(r_t^{\mathrm{app}}+r_t^{\mathrm{goal}}+r_t^{\mathrm{bear}})\\
&-0.5(1-v_{t+1})\min\{\tau_{t+1}/(0.5\,\mathrm{s}),1\}\\
&-0.01\lVert\vect{a}_t^{\mathrm{exec}}\rVert^2-0.001
+10\mathbf{1}[G_{t+1}]+r_t^{\mathrm{term}}.
\end{aligned}
\label{eq:task_reward}
\end{equation}
Here, $r_t^{\mathrm{app}}$ rewards reductions in robot-to-ball distance, $r_t^{\mathrm{goal}}$ rewards reductions in ball-to-goal distance, and $r_t^{\mathrm{bear}}$ rewards facing the ball. The $r_t^{\mathrm{app}}$ and $r_t^{\mathrm{bear}}$ terms are clipped to limit the effect of noisy estimates. Following a successful dribble, $r_t^{\mathrm{app}}$ is suppressed when the ball simultaneously advances toward the goal, avoiding a penalty from the resulting increase in robot-to-ball distance.

When the ball is not freshly observed, a separate penalty increases with the age of the most recent observation and saturates at $0.5\,\mathrm{s}$. The reward additionally includes an action-magnitude cost, a per-step cost, a $+10$ goal-crossing bonus, and a terminal reward. $G_{t+1}$ denotes the goal-crossing condition evaluated from the estimated ball position at $t+1$. Because this estimate can produce false positives, physical goal crossings are evaluated separately from the training reward.

\subsection{Termination \& Episode Resets}

An episode terminates upon scoring a goal, exceeding field boundaries, losing visibility of the ball for an extended period of time, or reaching the episode step limit.
Invalid telemetry, communication failures, and falls also terminate the episode.
After termination, an operator manually resets the robot and ball and starts a new episode. No transitions are recorded during this period.

\section{RSSM--RLPD for On-Robot Learning}
\label{sec:method}
% In this section you also need to highlight that this is a main algorithmic contribution and you need to say why we introduce this architecture

This section introduces our main algorithmic contribution, \method{}, which addresses partial ball visibility by augmenting the replay-mixture approach of RLPD \cite{ball2023rlpd}
with an observation-gated recurrent state-space model. The actor and critics share a representation trained by prediction
losses and the critic objective to retain information about ball motion
relevant to the control task.

\subsection{Observation-Gated Belief}

Let $\vect h_t\in\mathbb R^{128}$ and $\vect z_t\in\mathbb R^{16}$ denote the deterministic and stochastic components of the RSSM latent state, respectively, and let $\theta$ denote the parameters of the RSSM. The deterministic state is updated by the gated recurrent unit (GRU)  
\begin{equation}
 \vect h_t=\operatorname{GRU}_\theta\!\big(\vect h_{t-1},
  [\vect z_{t-1},\vect a_{t-1},\Delta t,\vect c_t]\big),
 \label{eq:method-recurrence}
\end{equation}
where $\eta_\theta$ is a learned input projection, $\Delta t$ is the interval between observations $o_{t-1}$ and $o_t$, and $c_t=e_\theta([o_t^{\mathrm{robot}},v_t])$ is a learned encoding of the robot pose and the ball visibility flag.
% We use the eight ball entries $o_t^{\mathrm{ball}}$ from
% Section~\ref{sec:task_observation} and write its four robot-to-goal entries
% as $o_t^{\mathrm{ctx}}\equiv o_t^{\mathrm{robot}}$. The RSSM implements the
% history summary of Section~\ref{sec:task_system} with two latent components:
% $h_t\in\mathbb R^{128}$ denotes its deterministic memory in this section,
% and $z_t\in\mathbb R^{16}$ is a stochastic latent state.
% Let $\theta$ denote the filtering-core parameters and $[\cdot]$ concatenation.
% A context encoder $e_\theta$ maps pose information and the filter's ball
% freshness flag $v_t$ to $c_t=e_\theta([o_t^{\mathrm{ctx}},v_t])$.
% The gated recurrent unit updates memory as
% \begin{align}
%  h_t &= \operatorname{GRU}_\theta\!\left(h_{t-1},
%  \eta_\theta([z_{t-1},\vect{a}_{t-1}^{\mathrm{rec}},\Delta t_t,c_t])\right),
%  \label{eq:method-recurrence}
% \end{align}
% where $\eta_\theta$ is a learned input projection,
% $\vect{a}_{t-1}^{\mathrm{rec}}$ is the preceding recorded command, and
% $\Delta t_t$ is the interval between observations $t-1$ and $t$.

We update the belief distribution over the stochastic state by selecting between two diagonal-Gaussian heads: the posterior $q_\theta(\vect z_t\mid\vect h_t,\vect o_t^{\text{ball}})$ when the ball is observed or the prior $p_\theta(\vect z_t\mid\vect h_t)$ when the ball is not observed. Ball observations therefore enter the belief through the posterior, while $c_t$ keeps the prior consistent with the robot's motion when the ball is out of view.
Let the superscript $s$ denote the selected prior or posterior. The actor and critics receive the belief features
\begin{equation}
 \vect b_t=\big[\vect h_t,\vect\mu^s_t,\log\vect\sigma^s_t,\
  v_t(\vect\mu^q_t-\vect\mu^p_t)\big]\in\mathbb R^{176},
 \label{eq:method-feature}
\end{equation}
where $\vect\mu^p_t$, $\vect\mu^q_t$, and $\vect\sigma^s_t$ are the means of the prior, posterior, and selected distributions, respectively, and $\vect\sigma^s_t$ is the element-wise standard deviation of the selected distribution. During training, $z_t$ is sampled from the selected distribution while actor and critic updates use its mean; the actor samples actions for exploration.

% The final block represents the amount in which the current observation moves the belief and is zero when the ball is not observed.
% Write $(\mu_t^p,\sigma_t^p)$ and $(\mu_t^q,\sigma_t^q)$ for the means and
% elementwise standard deviations of these distributions; superscript $s$
% denotes the selected distribution. The actor and critics receive
% \begin{equation}
%  f_t=[h_t,\mu_t^s,\log\sigma_t^s,
%                   v_t(\mu_t^q-\mu_t^p)]\in\mathbb{R}^{176}.
%  \label{eq:method-feature}
% \end{equation}
% The last block is the observation-induced correction, or innovation; the
% third describes learned latent dispersion. These features encode the
% model's internal belief statistics. Predictive training samples $z_t$ from
% the selected Gaussian; control and temporal-difference (TD) updates use
% its mean. The acting policy samples actions for exploration.

\subsection{Predictive Supervision from Replay}

We retain the original replay observation $o_t^{\mathrm{rec}}$ and freshness
$v_t^{\mathrm{rec}}$ as supervision when augmentation hides evidence from
the filter. A decoder of $[h_t,z_t]$ reconstructs $o_t^{\mathrm{rec}}$;
a separate decoder outputs a two-dimensional Gaussian over the estimated
robot-relative ball position. A transition decoder also receives
$\vect{a}_t^{\mathrm{rec}}$ and $\Delta t_{t+1}$ to predict interaction
outcomes. The model objective is
\begin{align}
 \mathcal L_M={}&\mathcal L_{\mathrm{obs}}+\mathcal L_r+
 \mathcal L_d+\mathcal L_v+\mathcal L_c\nonumber\\
 &+\mathcal L_{b,0}+\mathcal L_{b,+}
 +\mathcal L_K+\mathcal L_{K,+}.
 \label{eq:method-model-loss}
\end{align}
Terms have unit outer weights and average valid samples. MSE denotes
mean squared error, BCE binary cross-entropy, and NLL negative log-likelihood.

\smallskip
{\small\noindent
\begin{tabular}{@{}lp{0.80\columnwidth}@{}}
\toprule
Term & Supervision \\
\midrule
$\mathcal L_{\mathrm{obs}}$ & Recorded observation $o_t^{\mathrm{rec}}$ (MSE). \\
$\mathcal L_r$ & Recorded transition reward $r_t$ (MSE). \\
$\mathcal L_d$ & Transition termination-or-truncation label (BCE). \\
$\mathcal L_v$ & Next recorded freshness $v_{t+1}^{\mathrm{rec}}$ (BCE). \\
$\mathcal L_c$ & Weak contact from estimated motion (BCE). \\
$\mathcal L_{b,0}$ & Current recorded ball position (Gaussian NLL). \\
$\mathcal L_{b,+}$ & Future recorded ball position (Gaussian NLL). \\
$\mathcal L_K$ & Current posterior $q_t$ versus prior $p_t$ (KL). \\
$\mathcal L_{K,+}$ & Future posterior versus multistep prior (KL). \\
\bottomrule
\end{tabular}\par}
\smallskip

Reconstruction averages valid feature errors: robot-to-goal entries always contribute, and ball entries contribute when $v_t=1$. 
Ball-position NLL uses the first two entries of $o_t^{\mathrm{rec}}$ wherever $v_t^{\mathrm{rec}}=1$, including evidence hidden from the filter.
Reward, ending, and freshness losses average valid transitions; the window-final successor freshness target is zero. 
Weak contact labels use proximity, four-step estimated ball displacement, and action alignment. Their validity requires fresh evidence, suitable timing, and label permission; reacquisition and annotated external motion exclude affected labels.
Simulation supplies event annotations for this gating; hardware uses available perception and validity records. Ball targets are estimator measurements.

We regularize the posterior toward the prior with the balanced KL
\begin{align}
 \mathcal K(q,p)=\frac1{16}\sum_{j=1}^{16}\big[&
 \beta\max\{D_{\mathrm{KL}}(\operatorname{sg}(q_j)\Vert p_j),\epsilon\}
 \nonumber\\
 {}+&(1-\beta)\max\{D_{\mathrm{KL}}(q_j\Vert\operatorname{sg}(p_j)),
 \epsilon\}\big],
 \label{eq:method-balanced-kl}
\end{align}
where $j$ indexes latent dimensions, $\operatorname{sg}$ stops
gradients, $\beta=0.8$ weights the two gradient branches and $\epsilon=0.05$ is the per-dimension lower bound (free bits). $\mathcal L_K$ averages $\mathcal K(q_t,p_t)$ at fresh filter updates.

% For the KL terms, $D_{\mathrm{KL}}$ denotes Kullback--Leibler divergence,
% $q_j,p_j$ the one-dimensional latent marginals, and $\operatorname{sg}$ the
% stop-gradient operator, which preserves values while blocking gradients.
% We balance prior and posterior learning with
% \begin{align}
%  \mathcal K(q,p)=\frac1{16}\sum_{j=1}^{16}\big[&
%  \beta\max\{D_{\mathrm{KL}}(\operatorname{sg}(q_j)\Vert p_j),\epsilon\}
%  \nonumber\\
%  {}+&(1-\beta)\max\{D_{\mathrm{KL}}(q_j\Vert\operatorname{sg}(p_j)),
%  \epsilon\}\big],
%  \label{eq:method-balanced-kl}
% \end{align}
% where $\beta=0.8$ weights the two gradient branches and $\epsilon=0.05$
% is the per-dimension lower bound (free bits). $\mathcal L_K$ averages
% $\mathcal K(q_t,p_t)$ at fresh filter updates.

The $+$ terms average horizons $k\in\{1,3,5,10,20,40\}$ equally.
Starting from a sampled filtering state, prediction advances prior means
using recorded actions, intervals, context, and freshness. At $t+k$,
$\mathcal L_{b,+}$ scores the decoded ball Gaussian against the recorded
estimate, and $\mathcal L_{K,+}$ compares the multistep prior with the
filtering posterior at that same time. Each horizon averages valid
within-episode paths, using original visibility for ball targets and filter
visibility for KL. Ball NLL averages two coordinates; KL averages 16
latent dimensions. These predictions train the representation from replay.

\subsection{Critic-Guided Representation Learning}

Two action-value critics $Q_{\psi_i}$, with parameters $\psi_i$ for $i\in\{1,2\}$, train the RSSM to retain task-relevant information. The actor
$\pi_\phi$, parameterized by $\phi$, selects actions from $b_t$.
Overbars denote target quantities: a slowly updated target filter computes
successor features $\bar b_{t+1}$, and $Q_{\bar\psi_i}$ are target critics. With $a'\sim\pi_\phi(\cdot\mid\bar b_{t+1})$ and $d_t\in\{0,1\}$ indicating episode termination, the TD target and critic loss are
\begin{align}
 y_t &=r_t+\gamma\,(1-d_t)\min_i Q_{\bar\psi_i}(\bar b_{t+1},a'),\\
 \mathcal L_Q&=\mathbb E\left[\frac12\sum_{i=1}^{2}
 \big(Q_{\psi_i}(b_t,a_t^{\mathrm{rec}})
                     -\operatorname{sg}(y_t)\big)^2\right],
 \label{eq:method-td}
\end{align}
where the expectation is over sampled replay positions.

% For $a'\sim\pi_\phi(\cdot\mid\bar b_{t+1})$,
% the TD target $y_t$ and critic loss are
% \begin{align}
%  y_t &=r_t+\gamma m_t\min_i Q_{\bar\psi_i}(\bar b_{t+1},a'),\\
%  \mathcal L_Q&=\mathbb E\left[\frac12\sum_{i=1}^{2}
%  \big(Q_{\psi_i}(b_t,\vect{a}_t^{\mathrm{rec}})
%                      -\operatorname{sg}(y_t)\big)^2\right].
%  \label{eq:method-td}
% \end{align}
% Here $\gamma$ is the discount and the expectation averages over sampled replay
% positions. The scalar bootstrap mask is
% $m_t=1-\mathbf{1}[\mathrm{terminated}_t]$ for online data and retains its
% stored value for demonstrations. Thus, time-limit truncations bootstrap.
% TD targets use recorded rewards; the actor objective combines the mean
% critic value with entropy regularization and a learned temperature.

TD updates train the critics and RSSM; predictive updates train
the core and decoders. Actor and temperature updates treat $b_t$ as fixed
through $\operatorname{sg}(b_t)$. Sequential TD and predictive updates
share the core's Adam state. Core and decoder learning rates are $10^{-4}$;
the target filter follows each core update with Polyak coefficient $0.005$.

\subsection{Replay Masking and the Simulation Schedule}

Augmentation hides a contiguous 6--36-step fresh span in eligible replay
sequences by replacing ball position with the sentinel, zeroing velocity,
and clearing $v_t$. It retains recorded supervision, actions, rewards,
termination masks, and age-derived features. Predictive training thereby
uses recorded outcomes to supervise inference from incomplete inputs.

During simulation learning, we taper the augmented actor--critic sampling
distribution. For $n$ actor transitions received by the learner, define
\begin{equation}
 s(n)=\operatorname{clip}\!\left(\frac{n-20000}{10000},0,1\right),
 \quad p_S(n)=\tfrac12[1-s(n)].
 \label{eq:method-schedule}
\end{equation}
The clipping operation bounds the schedule progress $s(n)$ to $[0,1]$.
$p_S(n)$ is the requested fraction for each of the masked and event-centered
sequences in actor--critic training; predictive training keeps both at $1/2$.
The objectives share samples and masks through 20k, then draw independent
pools. From 30k, actor--critic training uses ordinary replay with its
natural observation loss. Hardware trials below 20k use the initial regime.
Event-centered windows cover annotated onsets, disturbances, or active
phases recorded from 20k onward. Other windows use length-weighted episode
selection and uniform starts. Eligibility and integer rounding determine
the realized sampling fractions.

Each pool contains 16 online and 16 demonstration windows of 64 observations.
The first 16 positions establish recurrent context, with gradients passing
through this prefix. Predictive losses use all valid positions; each critic
update selects 128 online and 128 demonstration positions after the prefix,
without replacement within its pool. A learner block performs eight critic
updates, one actor/temperature update alongside the eighth critic update,
and one model update. The joint update uses the same pre-update parameters;
model learning begins with actor--critic training. At episode boundaries
the actor clears its filter, and after parameter broadcasts it reprocesses
up to 256 retained history entries. Section~\ref{sec:experiments} gives the
synchronization settings and interaction budgets.

\section{Experiments}
\label{sec:experiments}

We evaluate our approach on the task of learning to dribble a ball into a goal through physical interaction. Our experiments combine hardware and simulation to leverage their complementary strengths: hardware captures real-world sensing noise and contact dynamics without simulation-induced approximations, while simulation enables multi-seed replication and systematic evaluation across a broader range of failure conditions.

\subsection{On-Robot Learning: Protocol and Results}
\label{sec:hardware}

\begin{figure}[h]
    \centering
    \includegraphics[width=0.7\linewidth]{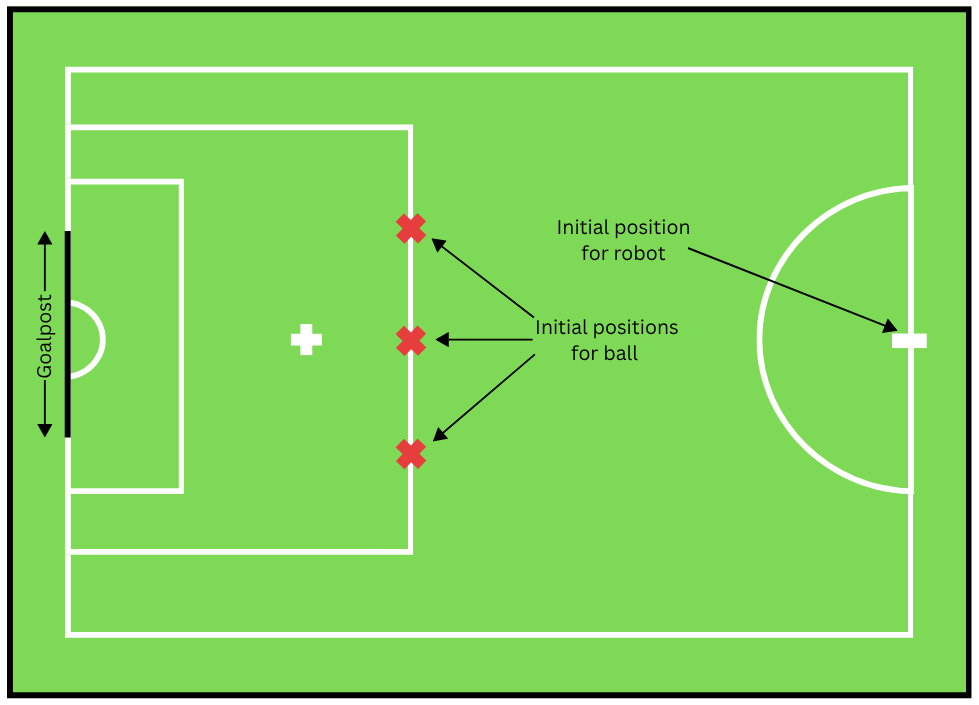}
    \caption{Outline of our real-world training environment setup}
    \label{fig:goal_location}
\end{figure}

\paragraph{Physical training protocol}
The physical task uses a $7\times4.8$~m half-field. Learner gradient updates run on an external workstation with an RTX 5090 GPU, which communicates wirelessly with the NAO robot. The robot starts at the center of the field, facing the target goal. The operator places the ball sequentially at three predefined penalty-box waypoints—center, left, and right—as shown in Figure \ref{fig:goal_location}. After a successful goal, the ball is moved to the next waypoint. Once all three waypoints are completed, the sequence resets to the center waypoint. Report first-goal timestamps, coverage of all three positions, subsequent
goals, and interventions. At budgets below 20k transitions, the 20--30k replay
adjustment in Section~\ref{sec:method} has not begun.
In all training runs, we provided each method with 50 expert demonstrations and limited the number of actions executed to 7000.

\begin{table}[h]
\centering
\caption{Real-robot training results}
\label{tab:hardware_training}
\footnotesize
\setlength{\tabcolsep}{2pt}
\begin{tabular}{lrrrr}
\toprule
Method & Goals & Time for First Goal & Actor Steps \\
\midrule
BC & 0 & N/A & 7,000 \\
Feed Forward    & 0 & N/A & 7,000 \\
GRU-RLPD           & 4 & 20 mins & 7,000 \\
\textbf{RSSM-RLPD}   & 9 & 10 mins & 7,000 \\
\bottomrule
\end{tabular}
\end{table}

\begin{figure}[h]
\centering
\definecolor{paperrssmrlpd}{HTML}{b12935}
\definecolor{papergrurlpd}{HTML}{1769a4}
\definecolor{paperffrlpd}{HTML}{d88700}
\definecolor{paperbc}{HTML}{646b73}

\begin{tikzpicture}
\begin{axis}[
    width=\columnwidth,
    height=4.65cm,
    scale only axis=false,
    xmin=0,
    xmax=65,
    ymin=0,
    ymax=9.5,
    xtick={0,10,20,30,40,50,60},
    ytick={0,1,2,3,4,5,6,7,8,9},
    xlabel={Elapsed time (minutes)},
    ylabel={Cumulative successes},
    label style={font=\footnotesize},
    tick label style={font=\scriptsize},
    axis lines*=left,
    ymajorgrids,
    grid style={gray!20},
    legend style={
        font=\scriptsize,
        draw=none,
        at={(0.5,1.05)},
        anchor=south,
        legend columns=2,
        /tikz/every even column/.append style={column sep=3pt}
    },
    legend cell align={left}
]

% RSSM-RLPD
\addplot[
    paperrssmrlpd,
    line width=0.85pt,
    mark=*,
    mark size=1.05pt,
    const plot
] coordinates {
    (0,0)
    (10,1)
    (14,2)
    (30,3)
    (34,4)
    (40,5)
    (42,6)
    (48,7)
    (50,8)
    (51,9)
    (62,9)
};
\addlegendentry{RSSM-RLPD}

% GRU-RLPD
% Run ended at 39 minutes.
\addplot[
    papergrurlpd,
    line width=0.85pt,
    mark=*,
    mark size=1.05pt,
    const plot
] coordinates {
    (0,0)
    (20,1)
    (24,2)
    (36,3)
    (39,4)
};
\addlegendentry{GRU-RLPD}

% FF-RLPD
\addplot[
    paperffrlpd,
    line width=0.85pt,
    dotted,
    mark=none
] coordinates {
    (0,0)
    (55,0)
};
\addlegendentry{FF-RLPD}

% Behavior Cloning
\addplot[
    paperbc,
    line width=0.85pt,
    dashed,
    mark=none
] coordinates {
    (0,0)
    (55,0)
};
\addlegendentry{BC}

\end{axis}
\end{tikzpicture}

\caption{Cumulative goal successes during real-robot training as a function of elapsed interaction time. The GRU-RLPD run terminates at 40 minutes, while the RSSM-RLPD run continues to 62 minutes, corresponding to the completion of 7,000 actor steps.}
\label{fig:training_success_time}
\end{figure}

\paragraph{Physical Robot Results}
% In all four of our training runs , we provided 50 expert demonstrations and limited the number of actions executed to 7000. 
As shown in Table \ref{tab:hardware_training}, \method{} scored nine goals compared to four goals from \grumethod{} and zero goals from both FF-RLPD and behavior cloning. \method{} was also able to score the first goal within 10 mins of training compared to 20 mins for \grumethod{}. Here goals are counts during the recorded session, and actor steps
denote rollout interactions. As shown in Figure \ref{fig:training_success_time} RSSM-RLPD achieves five goals within
40 minutes, while GRU-RLPD achieves four goals within 39 minutes.
FF-RLPD and behavior cloning do not achieve a goal during their respective
training runs.

% END HARDWARE UPDATE BLOCK

\begin{figure}[h]
    \centering
    \includegraphics[width=0.7\linewidth]{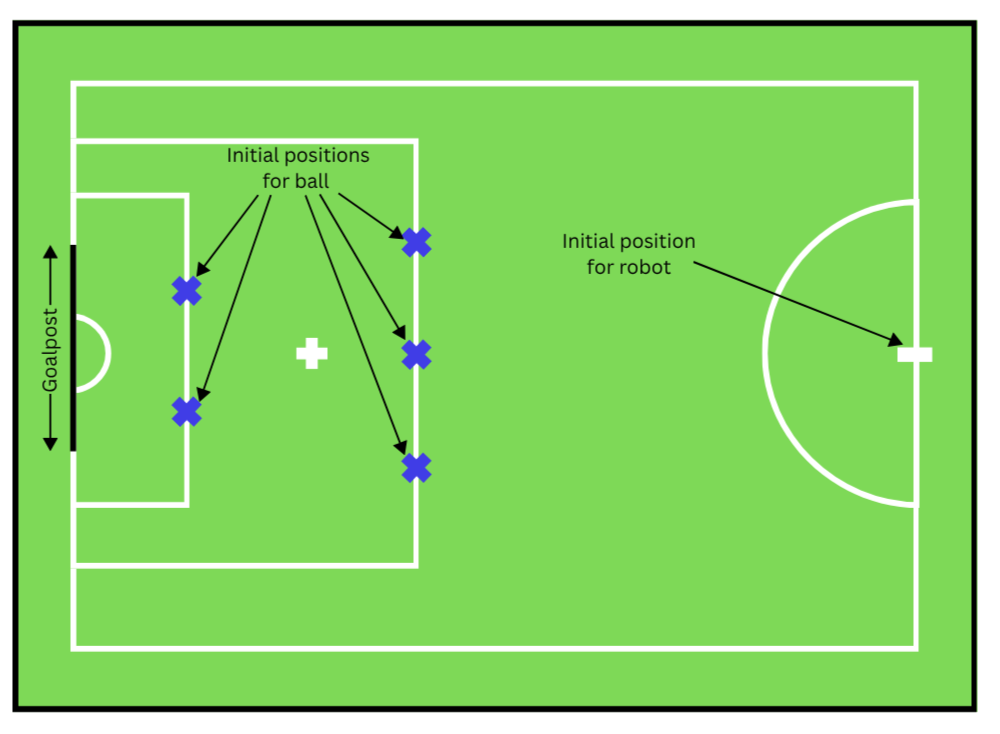}
    \caption{Outline of our real-world validation environment setup}
    \label{fig:val_goal_location}
\end{figure}

Figure \ref{fig:validation_success_time} shows the evaluation results for the checkpoints obtained from the on-robot training runs. The validation runs were designed to score 5 goals within 16 mins. In addition to the three points on the field that we used for training, there are 2 more points in the penalty box as shown in Figure \ref{fig:val_goal_location}. With \method{}, the robot scored all 5 goals within 10 mins and \grumethod{} within 16 mins. FF-RLPD scored a single goal, and behavior cloning scored none.  If we define success as scoring each goal within 4 mins, \method{} has a success of 100\%, \grumethod{} has a success of 80\%, and 0\% for both FF-RLPD and BC (Table~\ref{tab:validation_result}).

\begin{figure}[h]
\centering
\definecolor{paperrssmrlpd}{HTML}{b12935}
\definecolor{papergrurlpd}{HTML}{1769a4}
\definecolor{paperffrlpd}{HTML}{d88700}
\definecolor{paperbc}{HTML}{646b73}
\begin{tikzpicture}
\begin{axis}[
    width=\columnwidth,
    height=4.65cm,
    scale only axis=false,
    xmin=0,
    xmax=18,
    ymin=0,
    ymax=5.5,
    xtick={0,2,4,6,8,10,12,14,16,18},
    ytick={0,1,2,3,4,5},
    xlabel={Elapsed time (minutes)},
    ylabel={Cumulative successes},
    label style={font=\footnotesize},
    tick label style={font=\scriptsize},
    axis lines*=left,
    ymajorgrids,
    grid style={gray!20},
    legend style={
        font=\scriptsize,
        draw=none,
        at={(0.5,1.05)},
        anchor=south,
        legend columns=2,
        /tikz/every even column/.append style={column sep=3pt}
    },
    legend cell align={left}
]

% RSSM-RLPD
\addplot[
    paperrssmrlpd,
    line width=0.85pt,
    mark=*,
    mark size=1.05pt,
    const plot
] coordinates {
    (0,0)
    (1,1)
    (5,2)
    (8,3)
    (9,4)
    (10,5)
};
\addlegendentry{RSSM-RLPD}

% GRU-RLPD
\addplot[
    papergrurlpd,
    line width=0.85pt,
    mark=*,
    mark size=1.05pt,
    const plot
] coordinates {
    (0,0)
    (2,1)
    (11,2)
    (12,3)
    (15,4)
    (16,5)
};
\addlegendentry{GRU-RLPD}

% FF-RLPD
\addplot[
    paperffrlpd,
    line width=0.85pt,
    mark=*,
    mark size=1.05pt,
    const plot
] coordinates {
    (0,0)
    (8,1)
};
\addlegendentry{FF-RLPD}

% BC
\addplot[
    paperbc,
    line width=0.85pt,
    dashed,
    mark=none
] coordinates {
    (0,0)
    (18,0)
};
\addlegendentry{BC}

\end{axis}
\end{tikzpicture}

\caption{Cumulative goal successes during validation runs on the real robot, plotted against
elapsed time from the start of each run.}
\label{fig:validation_success_time}
\end{figure}

\begin{table}[h]
\centering
\caption{Real-robot validation results under 16 mins time budget.}
\label{tab:validation_result}
\small
\setlength{\tabcolsep}{5pt}
\begin{tabular}{lcc}
\toprule
Method & Success & Actions took\\
\midrule
BC & 0 & 16,722 \\
FF-RLPD & 0 & 5000 \\
GRU-RLPD & 80\% & 1900 \\
\textbf{RSSM-RLPD} & 100\% & 1500 \\
\bottomrule
\end{tabular}
\end{table}

\subsection{Simulation Empirical Setup}
\label{sec:sim_setup}
\paragraph{Task and training protocol}
We compare \method{}, \grumethod{}, feed-forward RLPD (FF-RLPD), and
behavior cloning (BC) using the same observation
(Section~\ref{sec:task_observation}). The simulator models sensing,
state estimation, walking, and ball contact.
We run five trials of each method. The three
online methods train from scratch for 100k transitions, mixing online and
demonstration samples equally. The shared archive contains 50 episodes collected on the real robot.

Training runs at 30 Hz with an 1800-step episode cap. Each termination
advances the center--left--right cycle in Table~\ref{tab:evaluation_setup}. The robot starts at $(0,0)$. Training resets use a stationary ball;
moving-ball validation starts at $0.20$~m/s. Evaluation fixes disturbances and physics; training randomizes them.
Instant resets restore poses, clear velocities and memory, and reinitialize
the estimator at the same simulation time. Physics is resampled per episode;
observation disturbances increase at 20k and 50k transitions. The first 500
transitions use random actions; learning starts after 1,000.

\paragraph{Baselines and optimization}
Actor and critic networks have three 256-unit tanh layers with layer
normalization, Adam learning rates of $3\times10^{-4}$, and discount 0.99.
Each learner block computes eight critic gradients and one actor/temperature
gradient. FF-RLPD uses batches of 256; recurrent methods use 32 windows of
length 64 with 16 TD burn-in positions.

FF-RLPD tests learning from the current estimator-derived observation alone. The strong \grumethod{} baseline
\cite{ni2022recurrent} uses a 176-unit TD-trained GRU, detached actor context,
and a target GRU. Its representation is trained by TD loss. It tests generic
recurrence against predictive memory at a matched representation width;
parameter counts differ. BC measures demonstration-only performance: it fits the same archive with 20,000 Adam
updates, the same actor architecture, batch size 256, and learning rate
$3\times10^{-4}$.

\paragraph{Validation protocol}
We evaluate actor checkpoints at $t=0, 10k, $\ldots$, 100k$
deterministically. Each training
placement location is crossed with nominal sensing and two settings of five perturbations: static-ball blackout, visible
kick followed by blackout, moving-ball burst loss, geometric occlusion, and
hidden kick during blackout. The resulting 33 scenes use shared sensor seeds 0 and 1:
66 attempts per trial and 330 per method/checkpoint. Evaluating BC once
gives 11,220 attempts overall. Rates pool attempts directly; nominal sensing
has fewer scenes than the other categories.

\begin{table}[t]
\centering
\caption{Shared training and validation placements, in field coordinates.}
\label{tab:evaluation_setup}
\small
\setlength{\tabcolsep}{5pt}
\begin{tabular}{lrrr}
\toprule
Placement & Ball $x$ (m) & Ball $y$ (m) & Robot yaw (rad) \\
\midrule
Center & 1.6 & 0 & 0 \\
Left & 1.6 & 1.0 & $+0.099$ \\
Right & 1.6 & $-1.0$ & $-0.099$ \\
\bottomrule
\end{tabular}
\end{table}

Each attempt covers one episode, ending at its first termination or
1800-step (60~s) limit. Training retains clipping, smoothing, safety shielding, and goal
keepout. Validation disables all action filters (including latency) and
goal-keepout termination/penalties for every frozen policy; tanh outputs
and dynamics are unchanged. Field/ball-out checks, 10-second continuous
ball loss, and scoring remain active. Center-line crossing is permitted during training and validation.
The suite measures performance at the three training placements.

\paragraph{Metrics and uncertainty}
The primary metric is \emph{episode success}: environment success
terminations, based on the fresh estimated ball position, divided by all
attempts. The episode success termination happens when the ball is detected to be in the goal reliably.  
Cumulative 20/40/60-second rates share
trajectories; early performance is trapezoidal 0--60k area under the success
curve (AUC), normalized by that interval.

\label{sec:sim_results}
\begin{table*}[t]
\centering
\caption{Simulation results under the 60-second protocol. Percentages have
pointwise 95\% trial-bootstrap intervals. AUC is normalized over 0--60k
transitions. Endpoints use 100k online policies or fixed BC, with 330 attempts
each; both success metrics use the same trajectories.}
\label{tab:main_results}
\small
\setlength{\tabcolsep}{4pt}
\begin{tabular}{lcrrr}
\toprule
Method & Online budget & Episode AUC$_{0\text{--}60k}$ & Mean episode success after 20k \\
\midrule
\method{} & 100k & 59.70 [57.35, 62.88] & 78.72 [73.30, 84.21] \\
\grumethod{} & 100k & 47.25 [36.82, 58.76] & 64.14 [53.16, 78.52] \\
FF-RLPD & 100k & 0.56 [0.00, 1.57] & 0.61 [0.13, 1.41] \\
BC & 0 & 3.94 [1.21, 7.58] & 3.94 [1.21, 7.58] \\
\bottomrule
\end{tabular}
\end{table*}

Pointwise 95\% percentile-bootstrap intervals use 10,000 resamples of the
five trials, retaining each trial's evaluation block and sharing draws
across methods and checkpoints. Uncertainty is conditional on the fixed
scene and sensor suite. Each interval describes a single checkpoint, with precision limited by
the five trials.

\subsection{Simulation Empirical Results}

\begin{figure}[h]
\centering
\includegraphics[width=\columnwidth]{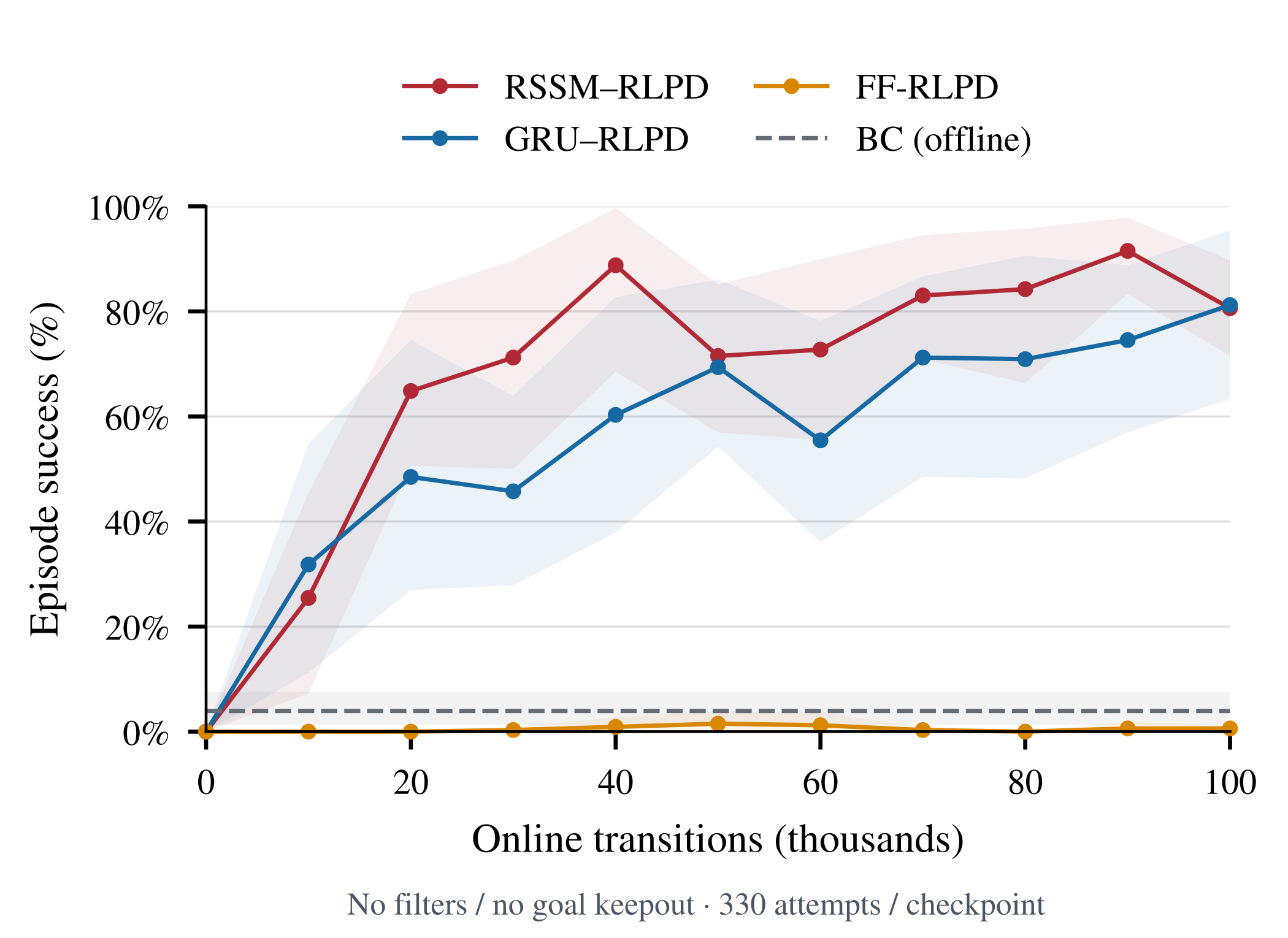}
\caption{Simulation episode success within 60~s over 330 attempts per
method/checkpoint. Bands show pointwise 95\% intervals from 10,000 paired
trial-bootstrap draws on the fixed evaluation suite. BC supplies the
constant offline reference.}
\label{fig:validation_learning}
\end{figure}

\paragraph{Learning from limited interaction}
Both recurrent methods substantially outperform FF-RLPD and BC
(Table~\ref{tab:main_results}, Fig.~\ref{fig:validation_learning}). FF-RLPD
remains below the offline BC reference at 100k. Under the same interaction
budget, the recurrent learners develop more successful policies from the
shared estimator input and demonstration. In simulation the control frequency is not aligned with physical robot experiment. So we focus on the overall performance tendency. 

\begin{table}[h]
\centering
\caption{Diagnostics at 100k (BC: offline), as percentages of 330 attempts.
Contact denotes any physical ball contact; ball loss denotes the 10-second
termination. }
\label{tab:interaction_diagnostics}
\small
\setlength{\tabcolsep}{3pt}
\begin{tabular}{lrrrr}
\toprule
Method & Contact & Ball loss & Timeout \\
\midrule
\method{} & 99.39 & 2.12 & 10.30 \\
\grumethod{} & 87.27 & 8.79 & 1.21 \\
FF-RLPD & 35.15 & 40.61 & 46.97 \\
BC & 32.73 & 85.15 & 3.94 \\
\bottomrule
\end{tabular}
\end{table}

\method{} improves learning over the strong \grumethod{} baseline: its mean
episode success is higher at every evaluated checkpoint from 20k to 90k,
and its normalized 0--60k success-curve area is larger. The paired AUC gain
is 12.45 percentage points (95\% interval [3.61,21.54]). This advantage
concerns learning efficiency: GRU leads at 10k, and the methods have similar
mean episode success at 100k. The observed advantage is concentrated in
the intermediate training interval.

\paragraph{Contact and failure modes}

\method{} makes contact more consistently and loses the ball less often
than the other methods, but times out more often than GRU
(Table~\ref{tab:interaction_diagnostics}). This pattern separates contact acquisition from timely completion. The RSSM
supplies memory while learning from the tracker estimates and recorded rewards.

\section{Limitations and Future Work}

In this section, we note some limitations of our study and discuss opportunities for future work.
Following RLPD \cite{ball2023rlpd}, our system used demonstration to make RL effective with limited interaction. We used a fixed demonstration dataset collected on the robot,
so learning may depend on its quality and coverage. Future work could
measure sensitivity to demonstration quality and quantity, and incorporate
targeted human corrections during learning.

In our work, we used an existing state estimator, whose errors can
affect both observations and rewards. Jointly improving estimation and
predictive memory could strengthen learning under noisy or missing
measurements.

Finally, the policy outputs three high-level walking commands, leaving foot placement and contact execution to the walking controller. This limits fine ball manipulation. Richer footstep or contact commands could enable more precise dribbling and a broader range of soccer skills but would also require more precise state estimation.

\section{Conclusion}
We presented RSSM-RLPD, a method for RL from real robot data that combines
prior demonstration data with an observation-gated recurrent state-space model for learning in tasks where it is difficult to estimate the robot's state using only on-board perception and localization.
In RSSM-RLPD, fresh ball observations correct a latent belief, while predictive updates maintain task-relevant memory during observation loss.
We apply RSSM-RLPD to the task of a NAO humanoid learning to dribble a ball into a goal using only onboard sensing. With this method the robot learned to score across multiple starting positions within 30 minutes of training time. In our hardware evaluation, RSSM-RLPD could score on all evaluation configurations within four minutes compared with none for
the basic RLPD approach under the same training budget.
In a simulation study, we found that a GRU-RLPD baseline also substantially outperformed basic RLPD and behavior cloning, highlighting the benefit of using recurrency to filter inaccurate state estimates. However, RSSM--RLPD achieved higher mean episode success than simply adding recurrency with a GRU.
% at every evaluated checkpoint from 20k to 90k transitions and a larger
% normalized 0--60k learning-curve area, with comparable performance at
% 100k transitions. 
These findings support observation-gated predictive
memory as a practical component of RL on physical platforms with limited state observability.
% ,
% enabling sustained ball interaction from intermittent, estimator-derived observations.

% \section*{Acknowledgment}

\bibliographystyle{IEEEtran}

\bibliography{curated}

\end{document}

%% file: figures/on_robot_loop_inline.tex
% One closed-loop system diagram; no external graphics are required.
\begin{figure}[h]
\centering
\begin{tikzpicture}[
  x=1cm,y=1cm,font=\scriptsize,
  block/.style={draw=black!65,rounded corners=2pt,align=center,
    inner sep=3pt,minimum height=0.56cm,fill=blue!4},
  signal/.style={-{Stealth[length=1.5mm]},line width=0.65pt,draw=black!80},
  learning/.style={-{Stealth[length=1.5mm]},dashed,line width=0.65pt,
    draw=green!35!black}
]
\path[use as bounding box] (0,-0.29) rectangle (8.05,3.89);

\node[block,fill=blue!10,text width=1.72cm,minimum height=0.85cm]
  (robot) at (1.0,2.15) {Physical robot\\Perception +\\state estimation};
\node[block,text width=1.43cm] (belief) at (3.60,3.00)
  {RSSM belief\\(offboard)};
\node[block,text width=1.23cm] (actor) at (6.55,3.00)
  {Actor\\(offboard)};
\node[block,text width=1.55cm] (walk) at (6.55,1.55)
  {Walking engine\\(onboard)};

\draw[signal] (robot.north) -- (1.0,3.00)
  -- node[above,inner sep=2pt] {estimates} (belief.west);
\draw[signal] (belief.east) -- node[above,inner sep=2pt] {$f_t$} (actor.west);
\draw[signal] (actor.south) -- node[right,align=left,inner sep=2pt]
  {filtered\\command} (walk.north);
\draw[signal] (walk.south) -- (6.55,1.05) -- (1.58,1.05) -- (1.58,1.725);

\node[block,fill=green!5,text width=1.40cm] (replay) at (1.0,0.32)
  {Online replay};
\node[block,fill=green!5,text width=1.60cm] (demos) at (3.17,0.32)
  {Demonstrations};
\node[block,fill=green!5,text width=1.96cm] (learner) at (6.05,0.32)
  {Offboard learner\\RLPD + RSSM};
\draw[learning] (robot.south) -- (replay.north);
\draw[learning] (replay.south) -- (1.0,-0.20)
  -- (6.05,-0.20) -- (learner.south);
\draw[learning] (demos.east) -- (learner.west);
\draw[learning,-] (learner.north) -- (4.73,0.83) -- (4.73,3.60);
\draw[learning] (4.73,3.60) -- (3.60,3.60) -- (belief.north);
\draw[learning] (4.73,3.60) -- node[above,inner sep=2pt] {new weights}
  (6.55,3.60) -- (actor.north);
\end{tikzpicture}
\caption{Physical interaction closes the control loop (solid arrows).
Online replay and demonstrations train the offboard learner, which updates
belief and actor parameters (dashed arrows). The RSSM augments onboard
estimation; the existing walking engine executes filtered commands.}
\label{fig:on_robot_loop}
\end{figure}